\documentclass[a4paper,11pt,final]{article}

\usepackage{epsfig}
\usepackage[cmex10]{amsmath}
\usepackage{amssymb,amsthm}
\usepackage{mathtools}
\usepackage{nccmath}
\usepackage{soul}
\usepackage[normalem]{ulem}
\usepackage{cite}
\usepackage{todonotes}
\usepackage{tikz}
\usepackage{hyperref}
\usetikzlibrary{hobby,decorations.markings,arrows,intersections,shapes}
\usepackage{pgfplots}
\usepgfplotslibrary{fillbetween}
\pgfplotsset{compat=newest}
\usepgfplotslibrary{fillbetween}
\usepackage{verbatim}

\begin{document}

\title{\ \\ \LARGE\bf Convergence analysis of particle swarm optimization using stochastic Lyapunov functions and quantifier elimination}

\author{Maximilian~Gerwien, Rick~Vo{\ss}winkel, and Hendrik~Richter \\
HTWK Leipzig University of Applied Sciences \\ Faculty of
Electrical Engineering and Information Technology\\
        Postfach 301166, D--04251 Leipzig, Germany. \\ Email: 
\{maximilian.gerwien,rick.vosswinkel,hendrik.richter\}@htwk-leipzig.de. }

\maketitle

\begin{abstract}

This paper adds to the discussion about theoretical aspects of particle swarm stability by proposing to employ stochastic Lyapunov functions and to determine the convergence set by quantifier elimination. We present a computational procedure and show that this approach leads to reevaluation and  extension of previously know stability regions for PSO using a Lyapunov approach under stagnation assumptions.

\end{abstract}

\section{Introduction}
Particle 
 swarm optimization (PSO) is a nature-inspired computational model that was originally developed by Kennedy and Eberhart~\cite{kennedy1995} to simulate the kinetics of birds in a flock.  Meanwhile,
PSO developed into a class of widely used bio-inspired optimization algorithms and thus the question of theoretical results about stability and convergence became important~\cite{vanderbergh2006,liu2015,Bonyadi2016a,Bonyadi2016b,cleghorn2015,cleghorn2018}.
Generally speaking, a PSO defines each particle as a potential solution to an optimization problem with a d-dimensional objective function $f:\mathbb{R}^d \rightarrow \mathbb{R}$. The PSO depends on three parameters, the inertial, the cognitive and the social weight. Stability analysis of PSO is mainly motivated by finding which combination of these parameters promotes convergence. 

In essence, PSO defines a stochastic discrete-time dynamical system. The stochasticity may be regarded as to come from two related sources. The primary source is that the prefactors of the cognitive and the social weight are realizations of a random variable, usually with a uniform distribution on the unit interval. A secondary source is that the sequences of local and global best positions are also effected by random as they reflect the search dynamics of the PSO caused by the interplay between the primary source of random and the objective function. Thus, these sequences can also be modeled as realizations of a random variable, but with a non-stationary distribution as the search dynamics modifies the influence of the primary source of random.   
 Most of the existing works on PSO stability focus on the primary source of random and assume stagnation in the sequences of personal and global best positions. 
Lately, there are some first attempts to incorporate the secondary source of random as well with stability analysis of a non-stagnate distribution assumption~\cite{Bonyadi2016a,cleghorn2018}. 

The stability of PSO implies that the sequence of particle positions remains bounded and some quantities calculated from this sequence converge to a target value. This has been done under the so-called deterministic assumption in which omitting random leads to deterministic stability analysis~\cite{vanderbergh2006,clerc2002particle,van2002analysis,trelea2003particle}.  If the random drive of the PSO is included in the analysis,  then a stochastic quantity on the sequence of particle positions could be the expected value, or variance, or even skewness and kurtosis, which leads to order-1~\cite{Poli2007,Poli2009},
order-2~\cite{Bonyadi2016a,liu2015,cleghorn2015,cleghorn2019}, and order-3~\cite{dong2019} stability.  
Basically, 
conditions for boundedness and convergence of PSO sequences can be obtained by different mathematical methods. The first group of methods focuses on the dynamical system's aspects of the PSO  and uses Lyapunov function arguments for stability analysis~\cite{vanderbergh2006,trelea2003particle,visakan_2006,Gazi2012}. Another group of methods explicitly addresses the stochastic character of the PSO and relies on non-homogenous recurrence relations~\cite{Bonyadi2016a,Poli2009} and convergence of a sequence of bounded linear operators~\cite{cleghorn2018}.

A stability analysis using Lyapunov function arguments is particularly interesting from a dynamical system's point of view as the method provides a mathematically direct way to derive stability conditions on system parameters. 
However, the method is also known to frequently giving rather conservative results and applications to PSO stability have indeed shown rather restrictive stability regions~\cite{trelea2003particle, visakan_2006}. 
To attenuate the conservatism of the Lyapunov approach to PSO stability, we propose in this paper an analysis consisting of two steps. First, we employ stochastic Lyapunov functions~\cite{semenov2003analysis,Gazi2012,li2013stability,correa2016} and subsequently determine the convergence set by quantifier elimination~\cite{roeb2018,roeb2019}. 
It is  shown that the convergence set we obtain by such a procedure gives a reevaluation and extension of previously know stability regions for PSO using a Lyapunov approach under stagnation assumptions.

The paper is organized as follows. In  Sec.~\ref{sec:pso}   we give a brief description of the PSO and discuss the mathematical description we use for the stability analysis. The stability analysis based on stochastic Lyapunov functions is given in Sec.~\ref{sec:lyap}, while Sec.~\ref{sec:conv} shows how the convergence set can be calculated by  quantifier elimination. Sec.~\ref{sec:results} gives the convergence region obtained by the computational procedure proposed in this paper, together with a comparison to existing results. 
The paper is concluded with a summary of the findings and some pointers to future work. 

\section{Particle Swarm Optimization} \label{sec:pso}
Let $\Gamma(t)$ be a set of $N$ particles in $\mathbb{R}^d$ at a discrete-time step $t$; we also call $\Gamma(t)$ the particle swarm at time $t$. Each particle is moving with an individual velocity and depends on a memory or feedback information about his own previous best position (local memory) and the best position of all particles (global memory). The individual next position at the next time-step $t+1$ results from the last individual position and the new velocity. Each particle is updated independently of others and the only link between the dimensions of the problem space is introduced via the objective function. Therefore, we can analyze the one-dimensional-one-particle case and keep the general case. 

The
general system equations \cite{vanderbergh2006,Poli2007,visakan_2006,engelb_2014} in the one-dimensional case are
\begin{align}
x(t+1) =& x(t) + v(t+1)\\
v(t+1) =&wv(t)+c_1r_1(t)p^l(t) - x(t)) \notag \\ &+c_2r_2(t)(p^g(t) - x(t)) \label{equ:diff_alg},
\end{align}
where $x(t)$ is the position and $v(t)$ is the velocity at time step $t$. The best local and global position is represented by $p^l$ and $p^g$; $w$ is the inertial weight. The random variables $r_1(t),r_2(t): \mathbb{R}^+ \rightarrow \mathcal{U}\left[ 0,1 \right] $ are uniformly distributed. The parameters $c_1$ and $c_2$ are known as cognitive and social weights and scale the interval of the uniform distribution, i.e. $c_1r_1(t) \sim \mathcal{U}\left[ 0,c_1 \right]$. From Eq.~\eqref{equ:diff_alg} we see that the system is stochastic and has two states. An equilibrium point can only be reached if $p^l(t) = p^g(t) = p(t)$ with the fix points $v^*= 0$,  $x^*=p(t)$.

Previous works 
analyzed the stability with the substitution
\begin{align}
\theta&= \theta^{(l)} + \theta^{(g)} = c_1r_1(t) + c_2r_2(t)\\\label{equ:theta}
p &= \frac{\theta^{(l)}p^{(l)} + \theta^{(g)}p^{(g)}}{\theta}
\end{align}
where $\theta$ is a constant~\cite{clerc2002particle,van2002analysis,trelea2003particle}. This is known as the deterministic assumption.
Thus, the simplified system definition in state-space form
\begin{equation}
\begin{pmatrix} 
x_{t+1} \\
v_{t+1}
\end{pmatrix}
=
\begin{pmatrix} 
1-\theta & w\\
-\theta & w
\end{pmatrix}
\begin{pmatrix} 
x_t \\
v_t
\end{pmatrix}
+
\begin{pmatrix} 
\theta \\
\theta
\end{pmatrix}p,\label{equ:SS_alg}
\end{equation}
can be reformulated as a linear time-invariant second-order system whose stability analysis is a straightforward application of linear theory.

In Kadirkamanathan et al.~\cite{visakan_2006} the system is reformulated as a linear time-invariant second-order system with nonlinear feedback, which relates to Lure's stability problem. 
The nonlinear feedback is described by the parameter $\theta, 0<\theta<c_1+c_2$. The global best value $p$ is considered as another state variable in the state vector $z=\begin{pmatrix}x_t-p & v_t\end{pmatrix}^T$. The assumption $p=p^{(l)}=p^{(g)}$ applies to the time-invariant case.
This results in the following linear state space model:
\begin{align}
\begin{pmatrix} 
x_{t+1}-p \\
v_{t+1}
\end{pmatrix}
=&
\begin{pmatrix} 
1 & w\\
0 & w
\end{pmatrix}
\begin{pmatrix} 
x_t-p \\
v_t
\end{pmatrix}
+
\begin{pmatrix} 
1 \\
1
\end{pmatrix}u_t \label{equ:SS_Kardi_1} \\
y_t = & \begin{pmatrix}1 & 1\end{pmatrix} \begin{pmatrix} 
x_t-p \\
v_t
\end{pmatrix}\\ 
u_t = & -\theta y_t. \label{equ:SS_Kardi_3}
\end{align}
Later, Gazi~\cite{Gazi2012} used the system definition~\eqref{equ:SS_Kardi_1}-\eqref{equ:SS_Kardi_3} and showed with a stochastic Lyapunov function approach the convergence region according to a  positive real argument for absolute stability following Tsypkin's result.

In the following stability analysis we adopt the description~\eqref{equ:SS_Kardi_1}-\eqref{equ:SS_Kardi_3} of the PSO as a linear time-discrete, time-invariant stochastic system and reformulate by
\begin{equation}
z(t+1)=A z(t) + B z(t) r(t)  + C u(t) r(t)\label{equ:sdgl},
\end{equation}
where $z=\begin{pmatrix}x&v\end{pmatrix}^T$ is the state-vector with $n$ states, $A$ a $n\times n$ matrix for the deterministic and $B$ a $n\times n$ matrix for the stochastic part of the state-space system. The system input $u(t)$ is formed by the input matrix $C$ and $r(t)$ is a uniformly distributed random variable of the stochastic system.

To avoid the loss of uniform distribution through a linear combination with different parameters $c_1,c_2$ of the random variables $r_1(t),r_2(t)$ in Eq.~\eqref{equ:theta}, the adjustment 
\begin{align}
\theta_1 &=\theta_1^l+\theta_1^g= cr_1(t)+cr_2(t) \label{equ:theta1}\\
\theta_2&=\theta_2^l+\theta_2^g= c_1r(t)+c_2r(t) \label{equ:theta2}
\end{align}
is used.
Under the assumptions~\eqref{equ:theta1} and~\eqref{equ:theta2} we can define two systems which both follow the structure of  Eq.~\eqref{equ:sdgl}: 
\begin{align}
\text{\textbf{System 1:}}\nonumber\\
\Sigma_1&=\begin{cases}
A_1=\begin{pmatrix}
1 & \omega \\ 0 & \omega
\end{pmatrix} \\
B_1=\begin{pmatrix}
-c & 0 \\ -c & 0
\end{pmatrix}\\
C_1=\begin{pmatrix}
c & c
\end{pmatrix} \\
r(t)= r_1(t)+r_2(t)\\
u(t)=\frac{\theta_1^{l}p^{l}(t) + \theta_1^{g}p^{g}(t)}{\theta_1}
\end{cases}
\label{equ:sys1_u}\\
\text{\textbf{System 2:}}\nonumber\\
\Sigma_2&=\begin{cases}
A_2=\begin{pmatrix}
1 & \omega \\ 0 & \omega
\end{pmatrix}\\
B_2=\begin{pmatrix}
-(c_1+c_2) & 0 \\ -(c_1+c_2) & 0
\end{pmatrix}\\
C_2=\begin{pmatrix}
(c_1+c_2) & (c_1+c_2)
\end{pmatrix}\\
r(t)= r(t)\\
u(t)=\frac{\theta_2^{l}p^{l}(t) + \theta_2^{g}p^{g}(t)}{\theta_2}.
\end{cases}
\label{equ:sys2_u}
\end{align}
For analyzing the PSO system~\eqref{equ:sdgl} the following expression is considered:
\begin{equation}
z(t+1) = A_i z(t)+ B_i z(t) r(t)\label{equ:sdgl_sys},
\end{equation}
where $u(t)=0$. We also adopt the stagnation assumption: $p^l(t) = p^l$,
and $p^g(t) = p^g$. This holds for all $t$ in the case of convergence.
During stagnation, it is assumed that each particle behaves independently. This means that each dimension is treated independently and the behavior of the particles can be analyzed in isolation.
The solution to the optimization problem in the origin is considered and can be shifted at any time with the help of a coordinate transformation.

This linear discrete-time time-invariant system with multiplicative noise was studied in~\cite{rami2002discrete, huang2008infinite} treating the linear-quadratic regulator problem and other control problems.
Previous literature~\cite{visakan_2006, Gazi2012} used the assumption~\eqref{equ:theta2} as a simplification of~\eqref{equ:theta1}. In the following we will show that the two systems with the assumptions~\eqref{equ:theta1} and~\eqref{equ:theta2} are not equivalent in the case of convergence.

\section{Lyapunov based stability analyses} \label{sec:lyap}
\subsection{Lyapunov methods in the sense of It\^{o}}
Lyapunov methods are very powerful tools from modern control theory used in analyzing dynamical systems. Many important results of deterministic differential equations have been generalized to stochastic It\^{o} processes \cite{arnold1974stochastic,mao2007stochastic,meyn2012markov} and Lyapunov methods to analyze stochastic differential equations developed from a theoretical side as well as in practical application~\cite{blythe2001stability,semenov2003analysis,correa2016}, e.g. by investigating the convergence of Neural Networks and Evolutionary Algorithms. In the case of time-discrete nonlinear stochastic systems, there exists a stability theory equivalent to the continuous case. Incipient with some linear stochastic time-discrete systems~\cite{dragan2006mean, rami2002discrete, huang2008infinite}, this goes over to nonlinear stochastic time-discrete systems~\cite{paternoster2000stability} and proposals of a general theory for nonlinear stochastic time-discrete systems~\cite{li2013stability} for different stability definitions. This enables the application of Lyapunov methods to arbitrary random distributions.

A general definition of discrete-time nonlinear stochastic systems is:
\begin{equation}
z(t+1) = f(z(t),r(t),t), \; \; z(t_0)=z_0, \label{equ:gen_sys}
\end{equation}
where $r(t)$ is a one-dimensional stochastic process defined on a complete probability space $(\Omega, F, Pr)$ and $z_0 \in R^n$ is a constant vector for any given initial value $z(t_0) = z_0 \in R^n$. It is assumed that $f(0,r(t),t)\equiv 0, \forall t \in \lbrace t_0 + k : k \in N^+ \rbrace$ with $N^+ :=\lbrace 1,2,\cdots \rbrace$, such that system~\eqref{equ:gen_sys} has the solution $z(t) \equiv 0$ for the initial value $z(t_0) = 0$. This solution is called the trivial solution or equilibrium point. As standing hypothesis we assume that $f(z(t),r(t),t)$ satisfies the local Lipschitz condition to ensure the existence and uniqueness of the solution.

There are various stability definitions for discrete-time nonlinear stochastic systems.

{\bf Definition 1: (Stochastic Stability)} {\it
We define that the trivial solution of system~\eqref{equ:gen_sys} is said to be stochastically stable or stable in probability if and only if for every $\epsilon>0$ and $h>0$, there exists $\delta = \delta(\epsilon,h,t_0)>0$, such that
\begin{equation}
Pr\lbrace\vert z(t) \vert < h\rbrace \geq 1 - \epsilon, \; \; t \geq t_0, \label{equ:stoch_stab}
\end{equation}
when $\vert z_0 \vert < \delta$. Otherwise, it is said to be stochastically unstable.}

{\bf Definition 2: (Asymptotic Stochastic Stability)} {\it
We say that the system~\eqref{equ:gen_sys} is asymptotically stochastically stable if the system is stochastically stable in probability according to Definition~\ref{def:stoch_stab}, and for every $\epsilon > 0$, $h>0$, there exists $\delta = \delta(\epsilon,h,t_0)>0$, such that
\begin{equation}
Pr\lbrace \lim\limits_{t \rightarrow \infty}{z(t)} = 0\rbrace \geq 1 - \epsilon, \; \; t \geq t_0, \label{equ:asym_stab_def}
\end{equation}
when $\vert z_0 \vert < \delta$.}

To find some mathematical criteria that satisfy these definitions,  a stability criterion of discrete-time stochastic systems is next given with a mathematical expectation of the probability mass function~\cite{li2013stability}.

{\bf Theorem 1:}
{\it The system~\eqref{equ:gen_sys} is stable in probability if there exist a positive-definite function $V(z(t))$, such that $V(0)=0$, $V(z(t))>0 \; \;  \forall z(t)\neq0$ and
\begin{equation}
E \lbrace \Delta V(z(t)) \rbrace \leq 0 \label{equ:stab_crit}
\end{equation}
for all $z(t) \in R^n$. The function $V$ is called a Lyapunov function.}

The proof of Theorem 1 (as well as the following Theorem 2 dealing with asymptotic stability) is shown in~\cite{li2013stability}.

{\bf Theorem 2:}
{\it The system~\eqref{equ:gen_sys} is asymptotically stable in probability if there exists a positive-definite function $V(z(t))$ and  a non-negative continuous strictly increasing function $\gamma(\cdot)$ from $R_+$ to $\infty$, with $\gamma(0)=0$ such that it vanishes only at zero and
\begin{equation}
E \lbrace \Delta V(z(t)) \rbrace \leq -\gamma(\vert z(t) \vert)  < 0 \label{equ:asym_stab_crit}
\end{equation}
for all $z(t) \in R^n$. The function V is called a Lyapunov function. }

\subsection{Applying the Lyapunov method to PSO}
Consider the quadratic Lyapunov function candidate
\begin{equation}
V=z(t)^TPz(t)
\label{eq:qlypc}
\end{equation}
with a real symmetric positive-definite $2 \times 2$ matrix
\begin{equation} \label{eq:p_matrix}
  P =\begin{pmatrix}
        p_1 & p_2 \\
        p_2 & p_3 \\
     \end{pmatrix}.
\end{equation}
This matrix is positive-definite if the elements of the matrix $\lbrace p_1,p_2,p_3\rbrace \in P$
ensue the Sylvester criterion~\cite{swamy1973sylvester}
\begin{align}
p_1 &> 0 \label{eq:sylv_one} \\
p_1p_3-p^2_2 &> 0, \label{eq:sylv_two}
\end{align}
where the elements in $P$ have significant impact on the stability and the convergence region.
According to Theorem 2, the system~\eqref{equ:sdgl_sys} is said to be asymptotically stable in probability if 
\begin{equation}
E\lbrace\Delta V(z(t))\rbrace < 0.
\end{equation}
Define the Lyapunov difference equation of system~\eqref{eq:qlypc} as
\begin{align}
\Delta V(z(t))=& V(z(t+1))-V(z(t)) \notag \\
=& \left(A z(t) + B z(t)\right)^T P \left(A z(t) + B z(t)\right)\notag \\
&- z(t)^TPz(t) \notag \\
=& z(t)^T(A^T P A + A^T P B \notag \\
&+ B^{T} P A + B^{T} P B -P)z(t), \label{equ:d_lyap}
\end{align}
which leads to
\begin{align}
E\lbrace\Delta V(z(t))\rbrace =&E\lbrace z(t)^T(A^T P A + A^T P B + B^{T} P A \notag \\
&+ B^{T} P B -P)  z(t)\rbrace < 0.
\label{eq:erwsys1}
\end{align}
Formally, we can describe the expectation of $\Delta V(z(t))$ with 
\begin{equation}
E\lbrace\Delta V(z(t))\rbrace=  \sum \sum \Delta V(x,v) \cdot Pr(x,v) 
\label{eq:sum_exp}
\end{equation}
but the sums cannot be calculated directly because we do not know the distribution of $Pr(x,v)$, which is non-stationary and changes for every time-step. Therefore, we approximate the expectation \eqref{eq:sum_exp} by calculating the expectation of the $k$th-moments of the known uniformly distributed random variable $r(t)\sim \mathcal{U}[0,1]$. This can be written as
\begin{equation}
    E\lbrace\Delta V(x,v)\rbrace= \Delta V(x,v\vert E\lbrace r\rbrace,E\lbrace r^2\rbrace, \cdots,E\lbrace r^k\rbrace).\label{}
\end{equation}
We thus get the expected value for $E\lbrace\Delta V(x,v)\rbrace$ by the expectation of the random variable of $E\lbrace r\rbrace$ and its $k$-th moments.
The rationale of this approximation is that by considering with $k=1,2$ the first and second moment we may obtain results comparable to results for order-1 and order-2 stability~\cite{Poli2009,Bonyadi2016a,liu2015,cleghorn2015,cleghorn2019}.

We applied a quadratic Lyapunov function candidate (\ref{eq:qlypc}) such that we calculate up to the second moment. This can be done if we consider 
\begin{align}
E\lbrace\Delta V(z(t))\rbrace= &E\lbrace V(z(t+1))\rbrace - V(z(t)) \notag \\
=& z(t)^T(A^T P A-P + A^T P B^* \notag \\
+& B^{^*T} P A + E\lbrace B^{T} P B \rbrace)z(t),  \label{eq:exp_gazi} 
\end{align}
where $B^*$ includes the first and $E\lbrace B^T P B \rbrace$ the second moment of the random variable $r(t)$. The order of the Lyapunov function candidate depends on the order of the moment for calculating the expectation of the random variable $r(t)$.
For $\Sigma_1$ \eqref{equ:sys1_u}, we can determine the expectation  
\begin{align}
B_1^*=E \lbrace B r(t) \rbrace & = \begin{pmatrix}
-c & 0 \\ -c & 0
\end{pmatrix} E \lbrace r(t) \rbrace \notag\\
& = \begin{pmatrix}
-1 & 0 \\ -1 & 0
\end{pmatrix}  E \lbrace cr(t) \rbrace
\end{align}
with
$E \lbrace cr(t) \rbrace = c(E \lbrace r_1(t) \rbrace +E \lbrace r_2(t) \rbrace) = c(\frac{1}{2}+ \frac{1}{2})=c$.
Furthermore, we calculate the expectation of the squared random variable $r(t)$ 
\begin{align}
E \lbrace B^T P B \rbrace & = \begin{pmatrix}
-c^2r^2(p_1+2p_2+p_3) & 0 \\ 0 & 0
\end{pmatrix}\notag\\
& = \begin{pmatrix}
p_1+2p_2+p_3 & 0 \\ 0 & 0
\end{pmatrix}  E \lbrace c^2r^2(t) \rbrace
\end{align}
with $E \lbrace c^2r^2(t) \rbrace = c^2(E \lbrace r_1(t) \rbrace +E \lbrace r_2(t) \rbrace)^2 = \frac{7}{6}c^2$. With these expectations we get for the difference equation of the Lyapunov candidate, Eq.~\eqref{equ:d_lyap}, and $\Sigma_1$:
\begin{align}
    E\lbrace\Delta V(z(t))\rbrace  =& \Delta V\left(z(t)\vert E\lbrace r(t)\rbrace, E\lbrace r^2(t) \rbrace\right) \notag\\
     = &v^2\left(-p_3 + \left(p_1+2p_2+p_3\right)w^2\right) \notag\\
    -& 2vx \Big(p_2 +w\big(\left(-1+c\right)p_1\notag\\
    +&\left(-1+2c\right)p_2+cp_3\big)\Big) \notag\\
    +& x^2 \frac{1}{6}c\big(\left(-12+7c\right)p_1+\left(-6+7c\right)2p_2\notag\\
    +& 7 c p_3\big).\label{equ:1}
\end{align}
For $\Sigma_2$ \eqref{equ:sys2_u}, we can determine the expectation with
\begin{align}
B_2^*=E \lbrace B r(t) \rbrace & = \begin{pmatrix}
-(c_1+c_2) & 0 \\ -(c_1+c_2) & 0
\end{pmatrix} E \lbrace r(t) \rbrace \notag\\
& = \begin{pmatrix}
-1 & 0 \\ -1 & 0
\end{pmatrix}  E \lbrace (c_1+c_2)r(t) \rbrace,
\end{align}
and $E \lbrace (c_1+c_2)r(t) \rbrace = (E \lbrace c_1r(t) \rbrace +E \lbrace c_2r(t) \rbrace) = \frac{(c_1+c_2)}{2}$. The expectation of the term $B^T P B$ can be expressed as follows
\begin{align}
E \lbrace B^T P B \rbrace & = \begin{pmatrix}
(c_1+c_2)^2r^2(p_1+2p_2+p_3) & 0 \\ 0 & 0
\end{pmatrix}\notag\\
& = \begin{pmatrix}
p_1+2p_2+p_3 & 0 \\ 0 & 0
\end{pmatrix}  E \lbrace (c_1+c_2)^2r^2 \rbrace. \label{equ:E(bpb)_sys2}
\end{align}
Furthermore, the expectation in the expression~\eqref{equ:E(bpb)_sys2} is calculated by
\begin{align}
    E \lbrace (c_1+c_2)^2r^2 \rbrace = & E \lbrace \left(\left(c_1+c_2\right) r \right)^2 \rbrace \notag \\
    = & \left( E \lbrace c_1 r\rbrace + E \lbrace c_2 r \rbrace \right)^2\notag \\
    = & E \lbrace c_1 r\rbrace^2 + 2E \lbrace c_1c_2 r\rbrace^2+E \lbrace c_2 r\rbrace^2 \notag \\
    = & \frac{1}{3}\left( c_1^2 + 2 c_1 c_2 + c_2^2 \right).
\end{align}
Now we can formulate the expectation of the difference equation of the Lyapunov candidate, Eq.~\eqref{equ:d_lyap},  
\begin{align}
    E\lbrace\Delta V(z(t))\rbrace  =& \Delta V\left(z(t)\vert E \lbrace (c_1+c_2)r(t) \rbrace, E \lbrace (c_1+c_2)^2r^2 \rbrace\right) \notag\\
     = &\frac{1}{3} \Big(3 v^2 ( - p_3 + ( p_1 + 2p_2 + p_3) w^2) \notag\\
    -& 3vx \big(2 p_2 + w( \left(-2 + c_1 + c_2\right) p_1 \notag\\
    +& 2\left( -1 + c_1 + c_2\right) p_2 + \left(c_1+c_2\right) p_3)\big)\notag\\
    +& x^2\left(c_1 + c_2 \right)\big( \left(-3 +c_1 +c_2\right) p_1 - 3 p_2 \notag\\
    +& \left( c_1 + c_2\right) \left(2 p_2 + p_3 \right)\big)\Big). \label{equ:2}
\end{align}

\section{Determining the convergence-set} \label{sec:conv}
Based on the Eqs.~\eqref{equ:1} and \eqref{equ:2} the convergence-set can be computed. To this end, we are looking for parameter constellations $(c_1+c_2)$ or $(c,\omega)$ for which the parameters $p_1,p_2,p_3$ specifying the quadratic Lyapunov function candidate (\ref{eq:qlypc}) exist such that the resulting matrix $P$, as defined by Eq. (\ref{eq:p_matrix}),  is positive definite and the condition $E\lbrace\Delta V(z(t))\rbrace <0$  holds in Eqs. \eqref{equ:1}  and \eqref{equ:2}. Technically, this can be expressed using existential ($\exists$) and universal ($\forall$) quantifiers, which gives
\begin{equation}
\exists p_1,p_2,p_3: E\lbrace\Delta V(z(t))\rbrace <0 \land p_1>0 \land p_1p_3-p_2^2>0,
\label{equ:1_prenex}
\end{equation}
again to apply for Eq.~\eqref{equ:1} as well as for Eq. \eqref{equ:2}. Unfortunately, this expression does not permit a constructive and algorithmic way for the convergence set analysis. We are rather looking for a description of the set without quantifiers.
A very powerful method to achieve such a description is the so-called quantifier elimination (QE). Before we continue with the determination of the convergence set using this technique, let us briefly introduce some necessary notions and definitions, see also~\cite{roeb2018,roeb2019}.

Eqs.~\eqref{equ:1} and \eqref{equ:2} can be generalized using the \emph{prenex} formula
\begin{equation}
\label{eq:prenex}
G(y,z):=(Q_1 y_1)\cdots (Q_l y_l)\,F(y,z),
\end{equation}
with $Q_i \in \left\lbrace \exists, \forall \right\rbrace $. The formula $F(y,z)$ is called \emph{quantifier-free} and consists of a Boolean combination of \emph{atomic} formulas
 \begin{equation}
 \varphi(y_1\dots,y_l,z_1,\dots,z_k) \,\tau\, 0,
  \end{equation}
    with $\tau \in \{=,\neq,<,>,\leq,\geq\}$. In our case we have the prenex formula~\eqref{equ:1_prenex}, the quantifier-free formulas $E\lbrace\Delta V(z(t))\rbrace <0  \land p_1>0 \land p_1p_3-p_2^2>0$, the atomic formulas $E\lbrace\Delta V(z(t))\rbrace <0$, $p_1>0$ and $p_1p_3-p_2^2>0$ as well as the \emph{quantified} variables $y=\{p_1,p_2,p_3\}$ and the \emph{free} variables $z=\{c_1+c_2,\omega\}$.  We are now interested in a quantifier-free expression $H(z)$ which only depends on the free variables.
  The following theorem states that such an expression always exists~\cite[pp. 69-70]{basu2006}.
  
{\bf Theorem 3: Quantifier elimination over the real closed field}
{\it For every prenex formula $G(y,z)$ there exists an equivalent quantifier-free formula $H(z)$.}

The existence of such a quantifier-free equivalent was first proved by Alfred Tarski~\cite{tarski1948decision}. He also proposed the first algorithm to realize such a quantifier elimination. Unfortunately, the computational load of this algorithm cannot be bounded by any stack of exponentials and thus it does not apply to non-trivial problems.
The first employable algorithm is the \emph{cylindrical algebraic decomposition} (CAD) \cite{collins1974quantifier}. This procedure contains four phases. First, the space is decomposed in semi-algebraic sets called cells. In every cell, we have for every polynomial in the quantifier-free formula $F(y,z)$ a constant sign. These cells are gradually projected from $\mathbb{R}^n$ to $\mathbb{R}^1$. These projections are cylindrical, which means that projections of two different cells are equivalent or disjoint. Furthermore, every cell of the partition is a connected semi-algebraic set and thus algebraic. Based on the achieved interval conditions  the prenex formula is evaluated in $\mathbb{R}^1$. The result is  afterward lifted to $\mathbb{R}^n$. This leads to the sought quantifier-free equivalent $H(z)$. Although the procedure has, in the worst case,  a doubly exponential complexity \cite{davenport1988}, CAD is the most common and most universal algorithm to perform QE.

The second prevalent strategy to solve QE problems is \emph{virtual substitution} \cite{weispfenning1988complexity,Loos1993,weispfenning1994}.  At the beginning the innermost quantifier of a given prenex formula is changed to $\exists y_i$ using $\forall y: F(y) \iff \neg(\exists y: \neg F(y))$. Based on so-called elimination sets a formula equivalent substitution is used to solve $\exists y: F(y)$. This is iterated until all quantifiers are eliminated. Virtual substitution can be applied to polynomials up to the order of three. Its computational complexity rises exponentially with the number of quantified variables.

The third frequently used strategy is based on the number of real roots in a given interval. Due to Sturm-Habicht sequences a \emph{real root classification} can be performed \cite{Gonzalez1989,Yang1996,Iwane2013}. This approach can lead to very effective QE algorithms. Especially for so-called sign definite conditions ($\forall y: y\geq 0 \implies f(y)>0$)  a very high performance is available. In this case, the complexity grows exponentially with just the degree of the considered polynomials.
While CAD results are simple output formulas,  the ones resulting of virtual substitution and real root classification are generally very complex and redundant. Hence, a subsequent simplification is needed.

During the last decades, some powerful tools to handle QE problems became available. The first tool which was applicable to non-trivial problems is the open source package QEPCAD (Quantifier Elimination by Partial Cylindrical Algebraic Decomposition)~\cite{collins1991}. 
The subsequent tool QEPCAD~B \cite{brown2003qepcad} is available in the repositories of all common Linux distributions.
The packages Reduce and Redlog are open-source as well and further contain virtual substitution based algorithms. This also holds for Mathematica. The library RegularChains~\cite{chen2014,chen2016quantifier} gives a quiet efficient CAD implementation for the computer-algebra-system Maple. Finally,  Maple provides the software package SyNRAC~\cite{Anai2003,YANAMI2007}, which contains CAD, virtual substitution and real root classification based algorithms.

\section{Computational results and comparison of different convergence-regions} \label{sec:results}
The following computational results are obtained by the Maple Package SyNRAC~\cite{Anai2003,YANAMI2007}.
 As mentioned before the used algorithms are very sensitive concerning the number of considered variables. Fortunately,  the chosen Lyapunov candidate function gives a possibility for reduction. From the Eqs.~\eqref{eq:sylv_one} and \eqref{eq:sylv_two} we can see that the elements of the principal diagonal need to be positive ($p_1, p_3 >0$). Furthermore, the conditions we are verifying are inequalities, cf. Eq.~\eqref{eq:erwsys1}. These inequalities can be scaled with some positive factor $\beta$ without loss of generality. We can describe this scaling by scaling the matrix $P$. Thus the matrix $\Tilde{P}=\beta P,\, \beta >0$ leads to the same results as the matrix $P$. With this property we can  normalized the matrix $P$ to an element of the principal diagonal, that is, we can set $p_1$ or $p_3$ to unity. This reduces the search space by one dimension.
 
Nevertheless, the computational complexity involved in the elimination process leads to the necessity of a further restriction of the search space. In fact, with the computational resources available to us, we could only handle three quantified variables. Since the quantified variables $x$ and $v$ need to be eliminated for useful statements, we are thus able to quantify one variable of $P=\left(\begin{smallmatrix} p_1 & p_2 \\ p_2 & p_3 \end{smallmatrix} \right)$, that is either $p_1, p_2$ or $p_3$. The variable $p_2$ occurs twice in the matrix $P$. This results in more complex prenex formulas, especially with more terms of high degree, concerning the quantified variables. Considering this limitations, we setup three candidates of $P$ for our experiments.

At first, we compute the quantifier-free formula with
\begin{equation}
    P^*_1=\mathbb{I},  \label{equ:cand_P_1}
\end{equation} with the identity matrix $\mathbb{I}$. The quantifiers to be eliminated are therefore limited to the universal quantifier $\forall x,v $. In addition, we were able to compute the quantifier-free formula with
\begin{equation}
      P^*_2 =\begin{pmatrix}
        p_1 & 0 \\
        0 & p_3 \\
     \end{pmatrix},  \label{equ:cand_P_2}
\end{equation}
which makes it necessary to eliminate additionally the existence quantifier $\exists p_1,p_3$ and 
\begin{equation}
      P^*_3 =\begin{pmatrix}
        1 & p_2 \\
        p_2 & 1 \\
     \end{pmatrix},  \label{equ:cand_P_3}
\end{equation}
with two instances of the existence quantifier $\exists p_2$. As discussed above, the three  matrices~\eqref{equ:cand_P_1}-\eqref{equ:cand_P_3} represent all possible candidate matrices $P$ with two free variables. This means $P=\beta \mathbb{I}$ in matrix~\eqref{equ:cand_P_1} or  $P=\left(\begin{smallmatrix} 1 & 0 \\ 0 & p_3 \end{smallmatrix} \right)$  in matrix~\eqref{equ:cand_P_2} or $P=\left(\begin{smallmatrix} \beta & p_2 \\ p_2 & \beta \end{smallmatrix} \right)$ in matrix~\eqref{equ:cand_P_3}, each  with $0 <\beta < \infty$, leads to the same results.

 For all further computations we study the convergence-set with $c > 0$ and $w \in (-1,1)$. 
Every convergence expression reported below adheres to these conditions.  
The convergence-set for $\Sigma_1$ using the conditions~\eqref{equ:1_prenex} yields for  $P^*_1$ the quantifier-free formula
\begin{fleqn}[0pt]
\begin{align}
&H(c,\omega)=\notag\\ 
&7c -6<0\land \notag\\ 
&2 c^2 w^2 - 3 w^2 - 7 c^2 + 6 c > 0 \label{equ:Sys1I}
\end{align}
\end{fleqn}
for $P^*_2$ there is
\begin{fleqn}[0pt]
\begin{align}
&H(c,\omega)=\notag\\
&c < \text{root}_i(49 c^2 + 48 w_i^2 c - 168 c + 24 w_i^4 - 168 w_i^2 + 144) \label{equ:Sys1P1P3}
\end{align}
\end{fleqn}
and for $P^*_3$ the convergence-set is
\begin{fleqn}[0pt]
\begin{align}
&H(c,\omega)=\notag\\
&7c -6<0\land \notag\\
&7 c^2 - 24 w c - 6 c + 24 w^2 + 12 w - 12 < 0. 
  \label{equ:Sys1P2}
\end{align}
\end{fleqn}
 Calculating the convergence-set for $\Sigma_2$ using the condition~\eqref{equ:1_prenex} leads  for $P^*_1$ to the quantifier-free formula
\begin{fleqn}[0pt]
\begin{align}
&H(c_1+c_2,\omega)=\notag\\
&2(c_1 +c_2) -3\leq 0 \land \notag \\
&w^2(c_1 +c_2)^2-2(c_1 +c_2)^2+3(c_1 +c_2)-3w^2\geq 0,
\end{align}
\end{fleqn}
while for $P^*_2$ the convergence-set is
\begin{fleqn}[0pt]
\begin{align}
&H(c_1+c_2,\omega)=\notag \\
&3w^2+(c_1 +c_2) -3\leq 0 \land \notag \\
&3w^4+3w^2(c_1 +c_2)  \notag \\
&+(c_1 +c_2)^2-12w^2-6(c_1 +c_2)+9\geq 0\label{equ:Sys2P1P3},
\end{align}
\end{fleqn}
and for $P^*_3$ we have
\begin{fleqn}[0pt]
\begin{align}
&H(c_1+c_2,\omega)=\notag \\
&2 (c_1+c_2) - 3 \leq 0 \land\notag \\
&2 (c_1+c_2)^2 - 12 w (c_1+c_2) \notag \\
&- 3 (c_1+c_2) + 24 w^2 + 12 w - 12 \leq 0.
\label{equ:Sys2P2}
\end{align}
\end{fleqn}
\begin{figure}[h!]
\centering
\includegraphics[scale=1.2]{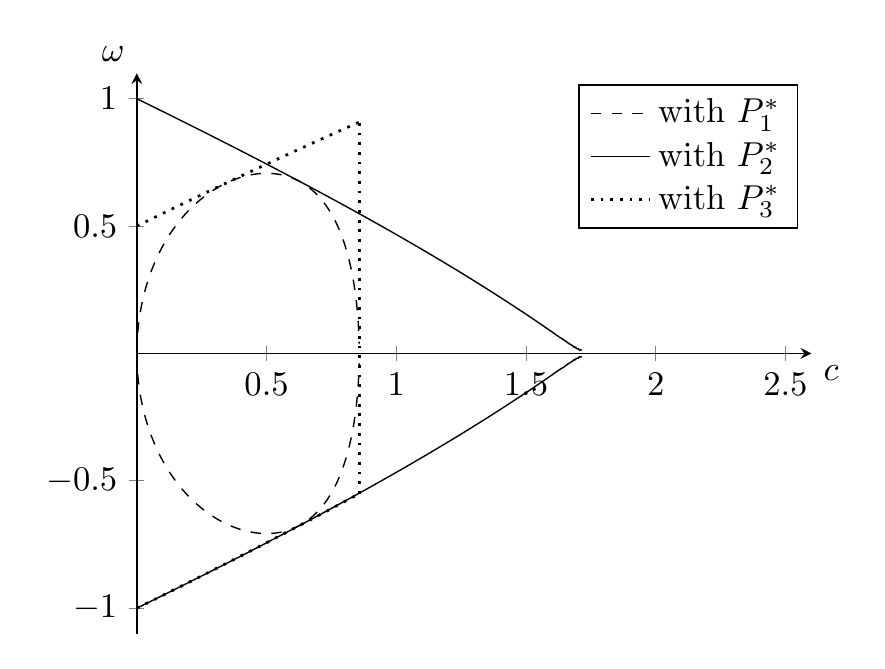}
\caption{Set of all parameter constellations assuring convergence in $\Sigma_1$.}
\label{fig:convergence-set_sys1} 
\end{figure}
\begin{figure}
\centering
\includegraphics[scale=1.2]{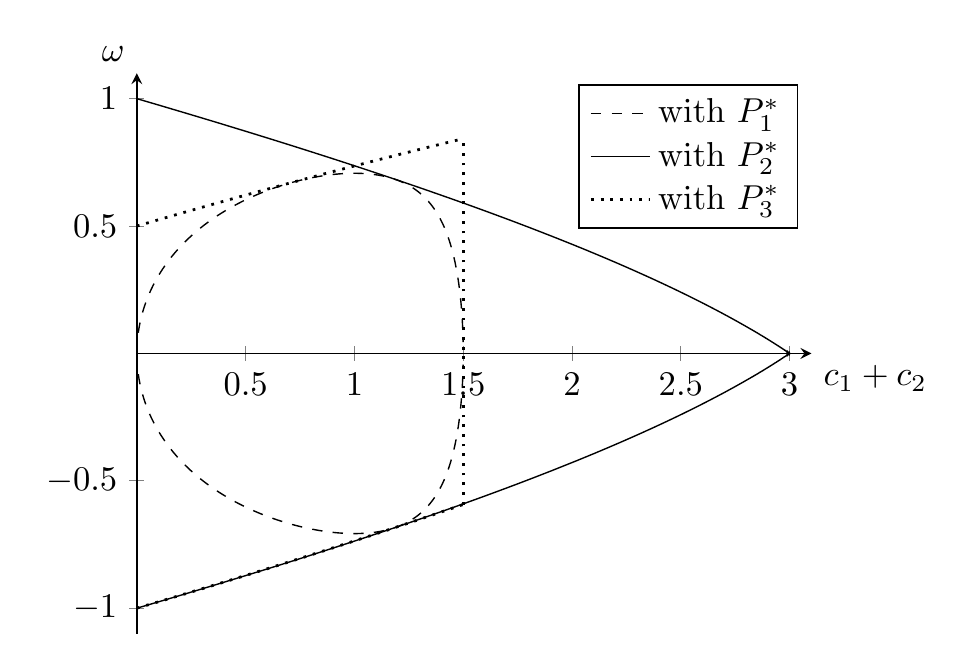}
\caption{Set of all parameter constellations assuring convergence in $\Sigma_2$.}
\label{fig:convergence-set_sys2} 
\end{figure}

The resulting set of parameters assuring convergence is illustrated in Fig.~\ref{fig:convergence-set_sys1} for $\Sigma_1$ and in Fig.~\ref{fig:convergence-set_sys2} for $\Sigma_2$. We can interpret these regions as the union of  possible solutions $\{p_1,p_2,p_3\} \in \mathbb{R}^3$, respecting the definiteness constraints, that is $p_1>0$ and $p_1p_3-p_2^2>0$. In other words, by considering $p_1$, $p_2$ and $p_3$ as quantified variables, we handle all numeric values at once, which includes the identity matrix. Thus, the result  for the identity matrix ($P_1^*$) gives a subset of $P_2^*$ as well as of $P^*_3$. With $P_2^*$  and $P_3^*$ either the quantifier $\exists p_1, p_3$  or $\exists p_2$  are eliminated.  Under the definiteness constraints we either get a solution for all possible $p_1$ and $p_3$ (but $p_2=0$) or all possible $p_2$ (but $p_1=p_3=1$). Thus,  $P_2^*$  and $P_3^*$ do not generalize each other and produce regions that partly overlap but also differ. If more powerful computational resources allow in the future to eliminate all three existence quantifiers at once, the resulting region should contain both the area for $P_2^*$  and $P_3^*$, thus generalizing the results given here.

\begin{figure}
\centering
\includegraphics[scale=1.2]{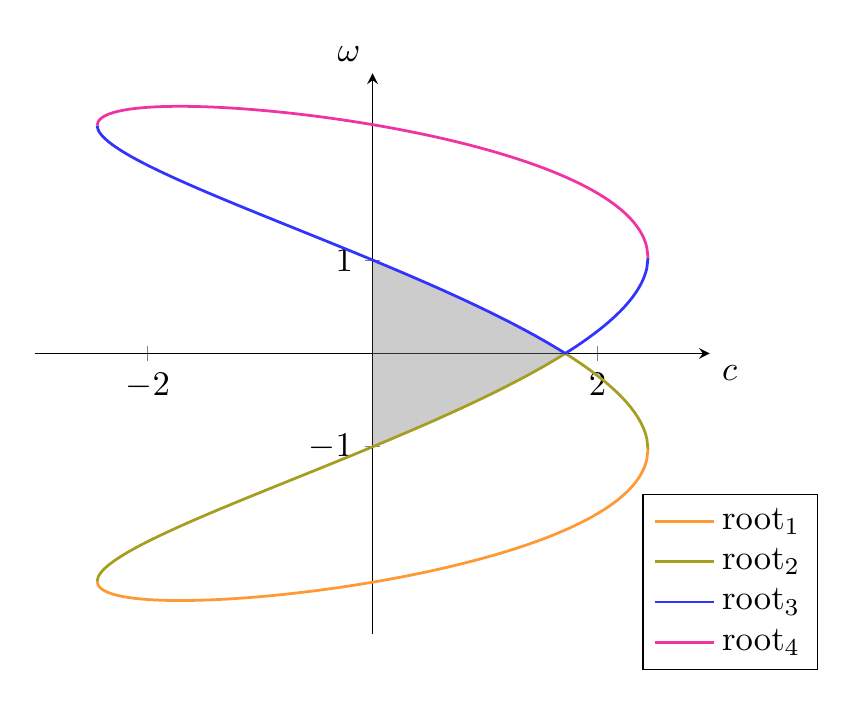}
\caption{Set of all roots of Expression~\eqref{equ:Sys1P1P3} for $\Sigma_1$ with convergence region.}
\label{fig:fancy_plot} 
\end{figure}

The expression~\eqref{equ:Sys1P1P3} is an indexed root expression. An indexed root expression $\text{root}_i$ can be written as
\begin{equation}
    \lbrace z_k \rbrace \;\tau \; \text{root}_i\varphi( z_1,\dots,z_k).
\end{equation}
It is true at a point $(\alpha_1, \cdots, \alpha_k) \in \mathbb{R}^k$. In our case the expression \eqref{equ:Sys1P1P3} are true for $ i=\lbrace2,3\rbrace$. In Fig.~\ref{fig:fancy_plot} we can see exemplary the set of all indexed root expressions $\text{root}_i$ of the convergence set~\eqref{equ:Sys1P1P3}. According to the constraints of $c>0$ and $w \in (-1,1)$, the solution of $\text{root}_2$ and $\text{root}_3$ bound the convergence region for $\Sigma_1$. 

We further observe that generally for $\Sigma_1$ and $\Sigma_2$ the convergence-region calculated with $P_1^*$ is also a subset of the regions calculated with $P_2^*$ and $P^*_3$. The area calculated for $\Sigma_2$  with $P_1^*$ leads to an expression with a quadratic polynomial and the region calculated with $P_2^*$ leads to an expression with a bi-quadratic polynomial, where the convergence-set is axisymmetric with respect to the $c=c_1+c_2$-axis. However, the derived region with $P^*_3$ possesses a point symmetry, for $\Sigma_1$ at point $\left(c,w \right)=\left(\frac{3}{7},0\right)$ and for $\Sigma_2$ at the point $\left(c_1+c_2,w \right)=\left(\frac{3}{4},0\right)$.
We can consider the convergence regions for the considered matrices $P$ in union. The union can be carried out because each region was derived under the same constraints and fulfills the stability condition according to Eqn.~\eqref{equ:1_prenex}, such that they do not contradict each other. In addition, the matrices~\eqref{equ:cand_P_2} and \eqref{equ:cand_P_3} each provided all possible instances of $p_1$ and $p_3$, or $p_2$, respectively. Thus, the non--overlapping regions calculated for $P^*_2$ and $P^*_3$ add to each other and thus results in a whole convergence-set.

\begin{figure}
\centering
\includegraphics[scale=1.0]{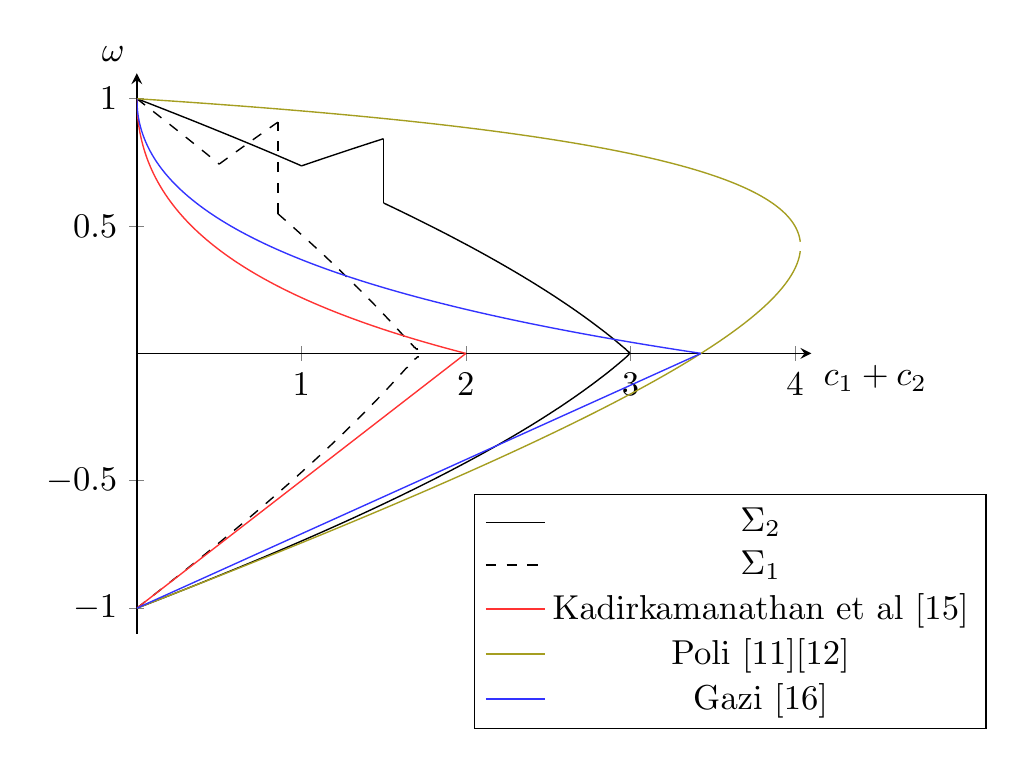}
\caption{Comparison of different theoretically derived convergence regions.}
\label{fig:convergence-set_compare} 
\end{figure}

We can finally set for $\Sigma_2$, $c_1+c_2 = c$ and compare $\Sigma_1$ and $\Sigma_2$. We can see that $\Sigma_1$ is nearly a subset of $\Sigma_2$. Only in the first quadrant a small triangle is not shared (see Fig.~\ref{fig:convergence-set_compare}). The reason for formulating two systems is avoiding the loss of uniform distribution through a linear combination in the system definition, see Eqn.~\eqref{equ:theta1} and \eqref{equ:theta2}. Maintaining the uniform distribution was important for calculating the expectation. Now we see that the simplification $c=c_1+c_2$ turns to another expectation and also another convergence set. 

In the past there were several approaches to derive the convergence region of particle swarms under different stagnation assumptions~\cite{Bonyadi2016a,Bonyadi2016b,cleghorn2015,cleghorn2018,engelb_2014}. In Fig.~\ref{fig:convergence-set_compare} we show our results as compared with other theoretically derived regions. Exemplary, we relate the results given by Eqn. \eqref{equ:Sys1I}-\eqref{equ:Sys2P2} to the findings of Kadirkamanathan et al.~\cite{visakan_2006}:
\begin{equation}
    H(c,w)=c<2(1+w) \land c < \frac{2(1-w)^2}{1+w} \label{equ:region_kardi}
\end{equation}
and Gazi~\cite{Gazi2012}:
\begin{equation}
H(c,w)= c< \frac{24 (1 - 2 \vert w \vert + w^2)}{7 (1 + w)}, \label{equ:region_gazi}
\end{equation}
with $w\in(-1,1)$.  Both stability regions are calculated according to the system definition~\eqref{equ:SS_Kardi_1}-\eqref{equ:SS_Kardi_3} and both are also based on using Lyapunov function approaches.
Kadirkamanathan et al.~\cite{visakan_2006} used a deterministic Lyapunov function and interpreted the stochastic drive of the PSO as
a nonlinear time-varying gain in the feedback path, thus solving  Lure's stability problem. Gazi~\cite{Gazi2012}  used a stochastic Lyapunov approach and calculated the convergence set by a positive real argument for absolute stability following Tsypkin's result.

Finally, we compare with the result of Poli~\cite{Poli2007} and \cite{Poli2009}:
\begin{equation}
    H(c,w)=c<\frac{24(1-w^2)}{7-5w},\label{equ:region_poli}
\end{equation}
which is calculated as the convergence region under the stagnation assumption but with the second moments of the PSO sampling distribution. The same convergence region~\eqref{equ:region_poli} has also been derived under  less restrictive stagnation assumptions, for instance for weak stagnation~\cite{liu2015} and  non-stagnate distributions~\cite{Bonyadi2016a,cleghorn2018}.   However, it was recently shown by Cleghorn~\cite{cleghorn2019} using numerical experiments with a multitude of different objective functions from the CEC 2014 problem set~\cite{cec2014_test_set} that the region expressed by Eq.~\eqref{equ:region_poli} at least for some functions slightly overestimates the numerical convergence properties of practical PSO. This was also suggested by some previous empirical results~\cite{engelb_2014}.

We now relate these theoretical results on convergence regions to each other. From Fig.~\ref{fig:convergence-set_compare} we can see that
the $\Sigma_2$ formulation proposed in this paper, expressions~\eqref{equ:Sys2P1P3} and \eqref{equ:Sys2P2}, contains the whole convergence region derived by Kadirkamanathan et al.~\cite{visakan_2006}, see Eq.~\eqref{equ:region_kardi}, and intersects with the region proposed by Gazi~\cite{Gazi2012}, see Eq. \eqref{equ:region_gazi}. Moreover, $\Sigma_2$  is a subset of inequality~\eqref{equ:region_poli}, which is the convergence-set first calculated by Poli and later re-derived by others under different stagnation assumptions~\cite{Poli2007,Poli2009,liu2015,Bonyadi2016a,cleghorn2018}. As calculating the region proposed by Gazi~\cite{Gazi2012} and our approach with $\Sigma_2$ both use stochastic Lyapunov functions, it may be interesting to compare them a little more. Comparing the blue (Gazi) and the black ($\Sigma_2$) curve in Fig.~\ref{fig:convergence-set_compare}, it can be seen that there is a substantial overlap, but also 
that $\Sigma_2$, especially by the derived convergence-set with $P^*_2$, is more bulbous and expand more in the first quadrant.
Under consideration of the derived region by the matrix $P^*_3$, we see one sub-region between $c\in(1,1.5)$ which expands the derived convergence-region computed by $P^*_2$ additionally with a small triangle. The region from Gazi is more elongated and extends longer for larger values of $c_1+c_2$ and small values around $w=0$. In terms of area, we have $A= 3.44 $ for $\Sigma_2$ and $A= 2.65 $ for Gazi.     
Both regions share about 98 percent (with respect to the Gazi-region), but  
the $\Sigma_2$ region is also about $23$ percent ($23.09 \%$) larger.

\section{Conclusion}
In this paper, 
we introduce a stochastic Lyapunov approach as a general method to analyze the stability of particle swarm optimization (PSO).  Lyapunov function arguments are interesting from a dynamical system's point of view as the method provides a mathematically direct way to derive stability conditions on system parameters. The method, however, is also known to provide rather conservative results. 
 We present a computational procedure that combines using a stochastic Lyapunov function with computing the convergence set by quantifier elimination (QE). Thus, we can show that the approach leads to a reevaluation and extension of previously know stability regions for PSO using Lyapunov approaches under stagnation assumptions. By calculating quantifier-free formulas with QE, we receive new analytical descriptions of the convergence region of two different system formulations of the PSO. The main difference to existing regions, apart from the size, is that the formulas describing them are more complex, with polynomial degrees up to quartic. 
 
The method presented offers to extend studying PSO stability even more. 
We approximated the distribution of $Pr(x,v)$ by the expectation of the random variable $E\lbrace r^k \rbrace$.  Poli~\cite{Poli2007,Poli2009}  proposed to calculate the so-called second-order moment of $Pr(v,x)$, which leads to rather good results.  The stochastic Lyapunov approach in connection with QE could also be applied to second-order moments of $Pr(x,v)$.  
We observed that there is a relation between the order of the Lyapunov candidate and the order of moments of the random variable. Thus, second-order moments can be accounted for by using Lyapunov function candidates of an order higher than quadratic. 

Another possible extension is to treat PSO 
 without stagnation assumptions~\cite{Bonyadi2016a,cleghorn2018}. This requires to consider the expected value of the sequences of local and global best positions, which results in another multiplicative term in the expectation of $\Delta V(x,v)$. Again, this could be accommodated by higher-order Lyapunov function candidates. 
The main problem currently is that higher-order Lyapunov function candidates also mean more variables that need to be eliminated by the QE. As of now, we could only handle, with the computational resources available to us, one quantified variable from the parameters of the Lyapunov function candidates, besides the system state $z=(x,y)$. This means an extension of the method proposed in this paper also relies upon further progress in computational hardware and efficient quantifier elimination implementations.





\section*{Supporting information}
We implemented our calculation of the convergence sets by stochastic Lyapunov functions and quantifier elimination in Maple programs using the package SyNRAC~\cite{Anai2003,YANAMI2007}. The visualization given in the figures was done in Mathematica. The code for both calculation and visualization is available at 
\url{https://github.com/sati-itas/pso_stabiliy_SLAQE}.



\begin{thebibliography}{10}

\bibitem{Anai2003}
H.~Anai and H.~Yanami.
\newblock {SyNRAC}: {A} {Maple}-package for solving real algebraic constraints.
\newblock In P.~M.~A. Sloot, D.~Abramson, A.~V. Bogdanov, J.~J. Dongarra, A.~Y.
  Zomaya, and Y.~E. Gorbachev, editors, {\em Computational Science --- ICCS
  2003: International Conference, Melbourne, Australia and St. Petersburg,
  Russia, June 2--4, 2003 Proceedings, Part I}, volume 2657 of {\em LNCS},
  pages 828--837. Springer, Berlin, Heidelberg, 2003.

\bibitem{arnold1974stochastic}
L.~Arnold.
\newblock {\em Stochastic Differential Equations}.
\newblock Springer, New York, 1974.

\bibitem{basu2006}
S.~Basu, R.~Pollack, and M.-F. Roy.
\newblock {\em Algorithms in Real Algebraic Geometry}.
\newblock Springer, Berlin, Heidelberg, 2 edition, 2006.

\bibitem{blythe2001stability}
S.~Blythe, X.~Mao, and X.~Liao.
\newblock Stability of stochastic delay neural networks.
\newblock {\em Journal of the Franklin Institute}, 338(4):481--495, 2001.

\bibitem{Bonyadi2016a}
M.~R. Bonyadi and Z.~Michalewicz.
\newblock Stability analysis of the particle swarm optimization without
  stagnation assumption.
\newblock {\em IEEE Transactions on Evolutionary Computation}, 20(5):814--819,
  2016.

\bibitem{Bonyadi2016b}
M.~R. Bonyadi and Z.~Michalewicz.
\newblock Particle swarm optimization for single objective continuous space
  problems: a review.
\newblock {\em IEEE Transactions on Evolutionary Computation}, 25(1):1--54,
  2017.

\bibitem{brown2003qepcad}
C.~W. Brown.
\newblock {QEPCAD B}: {A} program for computing with semi-algebraic sets using
  {CADs}.
\newblock {\em ACM SIGSAM Bulletin}, 37(4):97--108, 2003.

\bibitem{chen2014}
C.~Chen and M.~M. Maza.
\newblock Real quantifier elimination in the {Regular\-Chains} library.
\newblock In H.~Hong and C.~Yap, editors, {\em Mathematical Software --
  IMCS2014}, volume 8295 of {\em LNCS}, pages 283--290, Berlin, Heidelberg,
  2014. Springer-Verlag.

\bibitem{chen2016quantifier}
C.~Chen and M.~M. Maza.
\newblock Quantifier elimination by cylindrical algebraic decomposition based
  on regular chains.
\newblock {\em Journal of Symbolic Computation}, 75:74--93, 2016.

\bibitem{cleghorn2019}
C.~W. Cleghorn.
\newblock Particle swarm optimization: Understanding order-2 stability
  guarantees.
\newblock In P.~Kaufmann and P.~Castillo, editors, {\em Applications of
  Evolutionary Computation. EvoApplications 2019}, volume 11454 of {\em LNCS},
  pages 535--549. Springer, Berlin, Heidelberg, 2019.

\bibitem{engelb_2014}
C.~W. Cleghorn and A.~P. Engelbrecht.
\newblock Particle swarm convergence: An empirical investigation.
\newblock In {\em Proceedings of the 2014 IEEE Congress on Evolutionary
  Computation, CEC 2014}, 07 2014.

\bibitem{cleghorn2015}
C.~W. Cleghorn and A.~P. Engelbrecht.
\newblock Particle swarm variants: Standardized convergence analysis.
\newblock {\em Swarm Intelligence}, 9(2-3):177--203, 2015.

\bibitem{cleghorn2018}
C.~W. Cleghorn and A.~P. Engelbrecht.
\newblock Particle swarm stability: a theoretical extension using the
  non-stagnate distribution assumption.
\newblock {\em Swarm Intelligence}, 12(1):1--22, 2018.

\bibitem{clerc2002particle}
M.~Clerc and J.~Kennedy.
\newblock The particle swarm-explosion, stability, and convergence in a
  multidimensional complex space.
\newblock {\em IEEE Transactions on Evolutionary Computation}, 6(1):58--73,
  2002.

\bibitem{collins1974quantifier}
G.~E. Collins.
\newblock Quantifier elimination for real closed fields by cylindrical
  algebraic decomposition--preliminary report.
\newblock {\em ACM SIGSAM Bulletin}, 8(3):80--90, 1974.

\bibitem{collins1991}
G.~E. Collins and H.~Hong.
\newblock Partial cylindrical algebraic decomposition for quantifier
  elimination.
\newblock {\em Journal of Symbolic Computation}, 12(3):299--328, 1991.

\bibitem{correa2016}
C.~R. Correa, E.~F. Wanner, and C.~M. Fonseca.
\newblock Lyapunov design of a simple step-size adaptation strategy based on
  success.
\newblock In J.~Handl, E.~Hart, P.~R. Lewis, M.~Lopez-Ibanez, G.~Ochoa, and
  B.~Paechter, editors, {\em Parallel Problem Solving from Nature - PPSN XIV},
  volume 9921 of {\em LNCS}, pages 101--110. Springer, Berlin, Heidelberg,
  2016.

\bibitem{davenport1988}
J.~H. Davenport and J.~Heintz.
\newblock Real quantifier elimination is doubly exponential.
\newblock {\em Journal of Symbolic Computation}, 5(1):29--35, 1988.

\bibitem{dong2019}
W.~Y. Dong and R.~R. Zhang.
\newblock Order-3 stability analysis of particle swarm optimization.
\newblock {\em Information Sciences}, 503:508--520, 2019.

\bibitem{dragan2006mean}
V.~Dragan and T.~Morozan.
\newblock Mean square exponential stability for some stochastic linear discrete
  time systems.
\newblock {\em European Journal of Control}, 12(4):373--395, 2006.

\bibitem{Gazi2012}
V.~Gazi.
\newblock {Stochastic stability analysis of the particle dynamics in the {PSO}
  algorithm}.
\newblock {\em 2012 IEEE Multi-Conference on Systems and Control, MSC 2012},
  pages 708--713, 2012.

\bibitem{Gonzalez1989}
L.~Gonzalez-Vega, H.~Lombardi, T.~Recio, and M.-F. Roy.
\newblock Sturm-habicht sequence.
\newblock In {\em Proc. of the ACM-SIGSAM 1989 International Symposium on
  Symbolic and Algebraic Computation}, pages 136--146. ACM, 1989.

\bibitem{huang2008infinite}
Y.~Huang, W.~Zhang, and H.~Zhang.
\newblock Infinite horizon linear quadratic optimal control for discrete-time
  stochastic systems.
\newblock {\em Asian Journal of Control}, 10(5):608--615, 2008.

\bibitem{Iwane2013}
H.~Iwane, H.~Yanami, H.~Anai, and K.~Yokoyama.
\newblock An effective implementation of symbolic-numeric cylindrical algebraic
  decomposition for quantifier elimination.
\newblock {\em Theoretical Computer Science}, 479:43--69, 2013.

\bibitem{visakan_2006}
V.~Kadirkamanathan, K.~Selvarajah, and P.~J. Fleming.
\newblock Stability analysis of the particle dynamics in particle swarm
  optimizer.
\newblock {\em IEEE {T}ransactions on {E}volutionary {C}omputation},
  10(3):245--255, 2006.

\bibitem{kennedy1995}
J.~Kennedy and R.~C. Eberhart.
\newblock Particle swarm optimization.
\newblock {\em Proceedings of ICNN'95 - International Conference on Neural
  Networks}, pages 1942--1948, 1995.

\bibitem{li2013stability}
Y.~Li, W.~Zhang, and X.~Liu.
\newblock Stability of nonlinear stochastic discrete-time systems.
\newblock {\em Journal of Applied Mathematics}, 2013, 2013.

\bibitem{cec2014_test_set}
J.~Liang, B.~Y. Qu, and P.~N. Suganthan.
\newblock Problem definitions and evaluation criteria for the cec 2014 special
  session and competition on single objective real-parameter numerical
  optimization.
\newblock Technical report, Tech. Rep. 201311, Computational Intelligence
  Laboratory, Zhengzhou University and Nanyang Technological University, 2013.

\bibitem{liu2015}
Q.~Liu.
\newblock Order-2 stability analysis of particle swarm optimization.
\newblock {\em Evolutionary Computation}, 23(2):187--216, 2015.

\bibitem{Loos1993}
R.~Loos and V.~Weispfenning.
\newblock Applying linear quantifier elimination.
\newblock {\em The Computer Journal}, 36(5):450--462, 1993.

\bibitem{mao2007stochastic}
X.~Mao.
\newblock {\em Stochastic Differential Equations and Applications}.
\newblock Woodhead Publishing, Cambridge, 2007.

\bibitem{meyn2012markov}
S.~P. Meyn and R.~L. Tweedie.
\newblock {\em Markov Chains and Stochastic Stability}.
\newblock Springer Science \& Business Media, London, 2012.

\bibitem{paternoster2000stability}
B.~Paternoster and L.~Shaikhet.
\newblock About stability of nonlinear stochastic difference equations.
\newblock {\em Applied Mathematics Letters}, 13(5):27--32, 2000.

\bibitem{Poli2009}
R.~Poli.
\newblock Mean and variance of the sampling distribution of particle swarm
  optimizers during stagnation.
\newblock {\em IEEE Transactions on Evolutionary Computation}, 13(4):712--721,
  2009.

\bibitem{Poli2007}
R.~Poli and D.~Broomhead.
\newblock Exact analysis of the sampling distribution for the canonical
  particle swarm optimiser and its convergence during stagnation.
\newblock In {\em Proceedings of GECCO 2007: Genetic and Evolutionary
  Computation Conference}, pages 134--141, 2007.

\bibitem{rami2002discrete}
M.~A. Rami, X.~Chen, and X.~Zhou.
\newblock Discrete-time indefinite {LQ} control with state and control
  dependent noises.
\newblock {\em Journal of Global Optimization}, 23:245--265, 2002.

\bibitem{roeb2018}
K.~R{\"o}benack, R.~Vo{\ss}winkel, and H.~Richter.
\newblock Automatic generation of bounds for polynomial systems with
  application to the {L}orenz system.
\newblock {\em Chaos, Solitons \& Fractals}, 113:25--30, 2018.

\bibitem{roeb2019}
K.~R{\"o}benack, R.~Vo{\ss}winkel, and H.~Richter.
\newblock Calculating positive invariant sets: A quantifier elimination
  approach.
\newblock {\em Journal of Computational and Nonlinear Dynamics}, 14(7):074502,
  2019.

\bibitem{semenov2003analysis}
M.~A. Semenov and D.~A. Terkel.
\newblock Analysis of convergence of an evolutionary algorithm with
  self-adaptation using a stochastic {L}yapunov function.
\newblock {\em Evolutionary Computation}, 11(4):363--379, 2003.

\bibitem{swamy1973sylvester}
K.~Swamy.
\newblock On {S}ylvester's criterion for positive-semidefinite matrices.
\newblock {\em IEEE Transactions on Automatic Control}, 18(3):306--306, 1973.

\bibitem{tarski1948decision}
A.~Tarski.
\newblock {\em A Decision Method for a Elementary Algebra and Geometry}.
\newblock Project Rand. Rand Corporation, 1948.

\bibitem{trelea2003particle}
I.~C. Trelea.
\newblock The particle swarm optimization algorithm: convergence analysis and
  parameter selection.
\newblock {\em Information Processing Letters}, 85(6):317--325, 2003.

\bibitem{van2002analysis}
F.~van~den Bergh.
\newblock An analysis of particle swarm optimization.
\newblock {\em November, Ph. D. Dissertation, Faculty of Natural and
  Agricultural Sci., Univ. Pretoria, Pretoria, South Africa}, 2002.

\bibitem{vanderbergh2006}
F.~van~der Berg and A.~P. Engelbrecht.
\newblock A study of particle swarm optimization particle trajectories.
\newblock {\em Information Sciences}, 176:937--971, 2006.

\bibitem{weispfenning1988complexity}
V.~Weispfenning.
\newblock The complexity of linear problems in fields.
\newblock {\em Journal of Symbolic Computation}, 5(1-2):3--27, 1988.

\bibitem{weispfenning1994}
V.~Weispfenning.
\newblock Quantifier elimination for real algebra --- the cubic case.
\newblock In {\em Proc. of the International Symposium on Symbolic and
  Algebraic Computation}, ISSAC '94, pages 258--263. ACM, 1994.

\bibitem{YANAMI2007}
H.~Yanami and H.~Anai.
\newblock The {Maple} package {SyNRAC} and its application to robust control
  design.
\newblock {\em Future Generation Computer Systems}, 23(5):721--726, 2007.

\bibitem{Yang1996}
L.~Yang, X.~R.~Hou, and Z.~B.~Zeng.
\newblock Complete discrimination system for polynomials.
\newblock {\em Science in China Series E Technological Sciences}, 39(6), 1996.

\end{thebibliography}
\end{document}